\documentclass[10pt,twocolumn,letterpaper]{article}

\usepackage[accsupp]{axessibility} 
\usepackage{wacv}              

\usepackage{graphicx}
\usepackage{amsmath}
\usepackage{amssymb}
\usepackage{booktabs}

\usepackage{algorithm}
\usepackage{algorithmic}

\usepackage[title]{appendix}

\usepackage[pagebackref,breaklinks,colorlinks]{hyperref}

\usepackage[capitalize]{cleveref}
\crefname{section}{Sec.}{Secs.}
\Crefname{section}{Section}{Sections}
\Crefname{table}{Table}{Tables}
\crefname{table}{Tab.}{Tabs.}

\begin{document}

\title{Text Change Detection in Multilingual Documents Using Image Comparison}

\author{Doyoung Park, Naresh Reddy Yarram, Sunjin Kim, Minkyu Kim, Seongho Cho, Taehee Lee\\
Samsung SDS\\
{\tt\small \{dyj.park, reddy.naresh, sunjin07.kim, minkyu0828.kim, drizzle.cho, taehee23.lee\}@samsung.com}
}
\maketitle

\begin{abstract}
Document comparison typically relies on optical character recognition (OCR) as its core technology. 
However, OCR requires the selection of appropriate language models for each document and the performance of multilingual or hybrid models remains limited. 
To overcome these challenges, we propose text change detection (TCD) using an image comparison model tailored for multilingual documents. 
Unlike OCR-based approaches, our method employs word-level text image-to-image comparison to detect changes. 
Our model generates bidirectional change segmentation maps between the source and target documents. 
To enhance performance without requiring explicit text alignment or scaling preprocessing, we employ correlations among multi-scale attention features. 
We also construct a benchmark dataset comprising actual printed and scanned word pairs in various languages to evaluate our model. 
We validate our approach using our benchmark dataset and public benchmarks Distorted Document Images and the LRDE Document Binarization Dataset. 
We compare our model against state-of-the-art semantic segmentation and change detection models, as well as to conventional OCR-based models.
\end{abstract}

\section{Introduction}
\label{sec:intro}
In the digital age, the use of multilingual documents has proliferated, facilitated by globalization and the interconnectivity of international enterprises \cite{mukherjee2008impact01, bookbinder2009logistics02}. 
These documents often undergo various stages of editing, requiring precise tracking and confirmation of changes to the text across different versions.
Ensuring consistency and accuracy in these documents is critical for maintaining the integrity of the information they contain, especially in fields such as real estate, logistics, and finance.
Recently, there has been a trend towards storing these documents as images, such as scanned contract documents and digitized books.
This transition has presented new challenges and opportunities for document processing and analysis technologies.

One important strategy for image-based document comparison has been the use of optical character recognition (OCR).
Early OCR methods relied on pattern matching with standard character templates for character recognition \cite{tauschek1935reading03, schantz1982history04, mori1992historical05}. 
However, the emergence of deep learning (DL) techniques has led to significant improvements in various computer vision technologies, including classification, detection, segmentation, and OCR \cite{krizhevsky2017imagenet, simonyan2014very, girshick2014rich, breuel2013high06, anil2015convolutional07, lee2016recursive08}.
OCR technology is now widely used in a wide range of applications, including invoice processing, banking, legal documentation, and paper document digitization \cite{stolinski2011application09, singh2012survey10}.
The introduction of transformer models \cite{vaswani2017attention16} has also promoted innovative forms of document analysis, leading to further improvements in OCR systems \cite{huang2022layoutlmv17, fang2021read18}.
As a result, most off-the-shelf imaged document comparison software is based on OCR \cite{tafti2016ocr19}. 
This software integrates pre-processing steps, structural analysis, text detection, and various post-processing techniques to optimize OCR performance.
However, background noise, font variation, and multilingual text recognition can still affect OCR accuracy, particularly as the range of recognized languages increases.
In particular, handling multilingual documents places a significant burden on OCR systems, especially when encountering languages they have not been trained on or unexpected text formats.
Therefore, the present study proposes image comparison, which detects changes in an image or compares the quality between images, as a promising alternative to OCR for detecting text changes in multilingual documents.
By directly comparing text area images in order to bidirectionally detect changes between the source and target documents, our method eliminates the need for language-specific OCR models.

The major contributions of this study are as follows:
\begin{itemize}
    \item To the best of our knowledge, we propose the first text-image-based two-way change detection model that is independent of the language used in the text.
    \item We employ a correlation marginalization technique based on the surrounding features, reducing the need for extensive pre-processing steps.
    \item We construct a new imaged-text change-detection dataset and use it to demonstrate the state-of-the-art (SotA) performance of our proposed method, which is validated through ablation analysis.
\end{itemize}
Our model not only outperforms existing semantic segmentation and change detection methods but also achieves a performance comparable to traditional OCR approaches in document comparison tasks.

\section{Related Work}
\label{sec:related_works}
At its core, Document comparison involves textual analysis. 
Text comparison traditionally comprises two primary tasks: text detection and text recognition. 
However, after text detection is performed, our proposed method differs by using semantic segmentation for image-to-image comparisons rather than text recognition.
In particular, our approach focuses on detecting differences between documents by directly comparing images of text areas. 
When comparing text images, additional techniques are typically employed to clearly identify differences between documents, including noise removal and document layout analysis.
To explain the importance of our method, we briefly summarize relevant past studies in this section.

\subsection{Document image comparison}

An early study on document comparison introduced VisualDiff \cite{jain2013visualdiff}, which used SIFT \cite{lowe2004distinctive} to align and compare document images, demonstrating the potential for detecting document changes.
In another study that did not utilize OCR, Lin \etal \cite{lin2017fast} proposed a model that achieved fast speed and high precision in comparing multilingual documents.
However, this approach falls short of true multilingual functionality, relying solely on English, Chinese, and Japanese character features.
More recently, scanned document comparison based OCR has been studied \cite{andreeva2020comparison}, proposing a method that uses a combination of image comparison techniques to detect modifications between two versions of the same administrative document.
The proposed method establishes a basic document comparison process, but, OCR remains vulnerable to various limitations, such as low scan quality and multilingual documents. 
To address the challenge of multilingual comparison, a hierarchical document comparison method was recently proposed \cite{park2023document}, achieving superior performance in multilingual documents. 
However, this method is challenging to apply when there are numerous changes within the document due to the hierarchical structure of the method.
We propose a character image-based comparison method to address the problems of these previous studies. 
This method is language-independent, supports multiple languages, and is applicable even when many change, similar to OCR. 
Moreover, our method is more robust than OCR against image noise, such as from scanning.

\subsection{Text recognition}

Early text recognition methods relied on computer vision (CV) and machine learning techniques \cite{sarfraz2003offline, smith2007overview}.
Graves \etal proposed connectionist temporal classification (CTC), a method that uses a recurrent neural network (RNN) to decode features \cite{graves2006connectionist}. 
The performance of the technique has improved since the development of deep learning technology; however, a major drawback of these models is that they require a separate model for each language.
To address this issue, a multiplexed OCR model has been proposed which utilizes language prediction networks (LPNs) for language-adaptive recognition \cite{huang2021multiplexed}. 
However, this method introduces performance gaps between languages, and recent approaches must revert to specific training models for each language.
Since the development of deep learning technology, lightweight models such as PPOcr \cite{li2022pp} and transformer-based models such as TrOCR \cite{li2023trocr} have been proposed.
Although these methods support multilingualism, they are limited in comparing documents because each language require a separate model and distinct linguistic knowledge to be recognized.

\subsection{Semantic segmentation}

Semantic segmentation partitions images into distinct regions corresponding to specific objects or sections.
This technique has been used in applications such as autonomous driving and medical imaging.
Early approaches, such as fully convolutional networks (FCNs) \cite{long2015fully} and U-Net \cite{ronneberger2015u}, pioneered encoder-decoder architectures with skip connections to integrate features across various resolutions.
Sementic segmentation also shows strong performance in segmentation tasks, with transformer-based models such as Segformer \cite{xie2021segformer} proposed following the proliferation of transformers.
Transformer-based methods, such as BIT-CD \cite{chen2021remote} and SARAS-Net \cite{chen2023saras}, have recently gained attention as promising approaches for change detection using semantic segmentation, achieving SotA results in this area.
However, these change detection technologies are intended to find differences between images at different times at the same location.
While change detection technology excels at detecting meaningful changes, it is less effective at correcting positional errors and distortions introduced by document scanning or character detection.
\section{Method}

Text change detection (TCD) is utilized to identify altered text regions, such as additions, deletions, or modifications, by comparing text images extracted from two different versions of a document.
Our model has a two-way semantic segmentation architecture that analyzes text area image pairs from the source to the target and vice versa. 
We also use an auxiliary classification CNN head at the end of TCD network to determine if two text unit images are identical or not. 
Our model thus compares pairs of text unit images from the original and comparison documents at the character level to determine whether they are identical or different. 
Our model thus compares pairs of text unit images from the original and comparison documents at the character level to detect changes, additions, or deletions.
This section outlines the overall framework of our model and its key components.
Additionally, our overall document comparison application process is described in \cref{Appendix} to facilitate understanding of our proposed model and demonstrate its necessity.
\subsection{Overall architecture}

\begin{figure*}[t]
\centering
\includegraphics[width=0.9\textwidth]{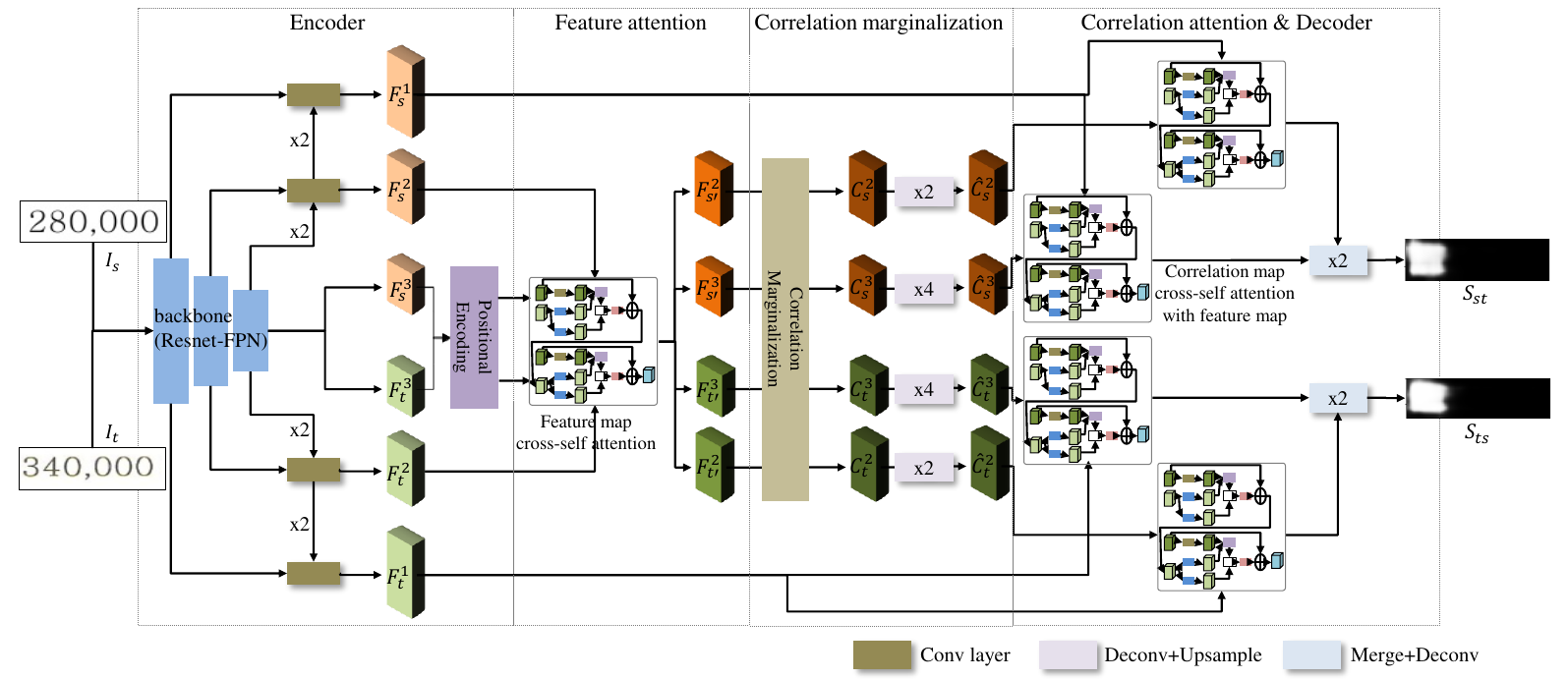}
\caption{Text image Change Detection Model Architecture. The architecture is composed of four modules: encoder, feature attention, correlation marginalization, and decoder, from left to right.}
\label{fig:TCD_architecture}
\end{figure*}

As shown in \cref{fig:TCD_architecture}, the architecture of our proposed model is based on an encoder–decoder structure \cite{badrinarayanan2017segnet}.
Both input images are processed using the encoder module with shared weights in a manner similar to Siamese networks \cite{chopra2005learning, koch2015siamese}, with ResNet \cite{he2016deep} used as the backbone. 
Given the diversity of text unit images, which include complex shapes and small characters such as math symbols or punctuation, multi-level features are extracted from ResNet using a feature pyramid network (FPN) \cite{lin2017feature} to preserve the fine details.
To ensure robust image comparison, our encoder employs cross-self attention mechanisms to enhance feature representation and highlight significant locations in the feature space.
After encoding, the features of the source and target images are separately fed into parallel decoders for each image, both of which generate change segmentation maps from the encoded features.
By utilizing features from the backbone and enhancing lower-resolution segmentation maps using the attention mechanisms, the decoders ensure the precise localization of changes to the text.

\subsection{Backbone network}
Given input images from the source $I_s$ and target $I_t$ of size $(H, W, 3)$, our ResNet backbone with the FPN generates three multi-scale feature pyramid map pairs of sizes $(H/2, W/2, 64)$, $(H/4, W/4, 64)$, and $(H/8, W/8, 512)$, denoted as $F_s^{1}, F_s^{2}, F_s^{3}$ and $F_t^{1}, F_t^{2}, F_t^{3}$ for the source and target, respectively. 
Our backbone is based on the ResNet-based network proposed by \cite{cheng2017focusing}, and we use only up to the third bottleneck layer.
The ResNet output is passed through the FPN to improve the feature maps, without changing the shape.

\subsection{Positional encoding and multi-scale feature map attention}
Inspired by transformer networks \cite{dosovitskiy2020image}, we employ a 2D extension of the standard positional encoding used in transformers similar to DETR \cite{carion2020end} and LoFTR \cite{sun2021loftr}.
We exclusively append these encodings to the FPN output feature map $F^{3}$ due to the limited spatial information present in the deeper features.
\label{multi_scale_feature_map_att}

Developing concepts from SARAS-Net \cite{chen2023saras} and LoFTR \cite{sun2021loftr}, our model incorporates cross-self attention modules, which utilize an attention learning module to enhance lower-level feature representation by focusing on meaningful features.
These modules utilize convolutional operations to derive queries, keys, and values, producing attention maps that refine feature representations \cref{eq:TCD_attention}. 
\begin{equation}
Attention(Q, K, V) = softmax(\frac{QK^T}{\sqrt{d_k}})V
\label{eq:TCD_attention}
\end{equation}
where ${Q}$, ${K}$, and ${V}$ denote the query, key, and value matrices respectively, and ${d_k}$ represents the dimension of the key matrix.
We employ feature map attention within each of the $F^2$ and $F^3$ feature maps.
At scales $(H/4, W/4, 64)$ and $(H/8, W/8, 512)$, the feature maps undergo the sequential application of cross-attention and self-attention; \eg, cross-attention is first computed with respect to feature map $F_s^2$ and self-attention is subsequently applied to refine the representation. 
Similarly, identical attention mechanisms are applied to $F_t^2$, $F_s^3$, and $F_t^3$, using the same protocol.

\subsection{Correlation map and marginalization}

We adopt the concept of hyper-correlation \cite{min2021hypercorrelation} in a few-shot segmentation area to construct a 4D correlation map using the cosine similarity between the multi-scale feature maps.
Specifically, the 4D correlation map between the two feature maps ${F_s^{l}}$ and ${F_t^{l}}$ at pixel positions ${(i,j)}$ and ${(m,n)}$, respectively, for ${l}$=2,3 is constructed via cosine similarity as follows:

\begin{small}
\begin{equation}
    {\pmb{Cos}(\pmb F_s(i,j), \pmb F_t(m,n)) = \pmb{ReLU}(\frac {\pmb F_s(i,j) \cdot \pmb F_t(m,n)}{||\pmb F_s(i,j)|| \cdot ||\pmb F_t(m,n)||})} 
\label{eq:TCD_cosine_simmilarity_4d_corr_map}
\end{equation}
\end{small}%

In text matching, we posit that the query points in the source and their corresponding points in the target coincide at analogous spatial locations.
This assumption is based on the fact that, when comparing text images, the target image may undergo transformations such as rotation, blurring, or the addition of whitespace or noise.
However, the same characters tend to appear in similar positions in both the source and target images.
Rather than computing a full, dense 4D correlation map, we confine the matching process to neighboring points surrounding the queried feature point, thus reducing the computational complexity.
The neighborhood is defined as [$-K_{v}, K_{v}$] and [$-K_{h}, K_{h}$] in the vertical and horizontal directions, respectively.
By calculating the sparse correlation within a predetermined range, we significantly reduce the computation time required compared to the full correlation map.
Correlation map marginalization refers to the process of transforming the 4D correlation tensor into two 3D correlation tensors, facilitating further analysis.

\begin{figure}[t]
\centering
\includegraphics[width=0.8\columnwidth]{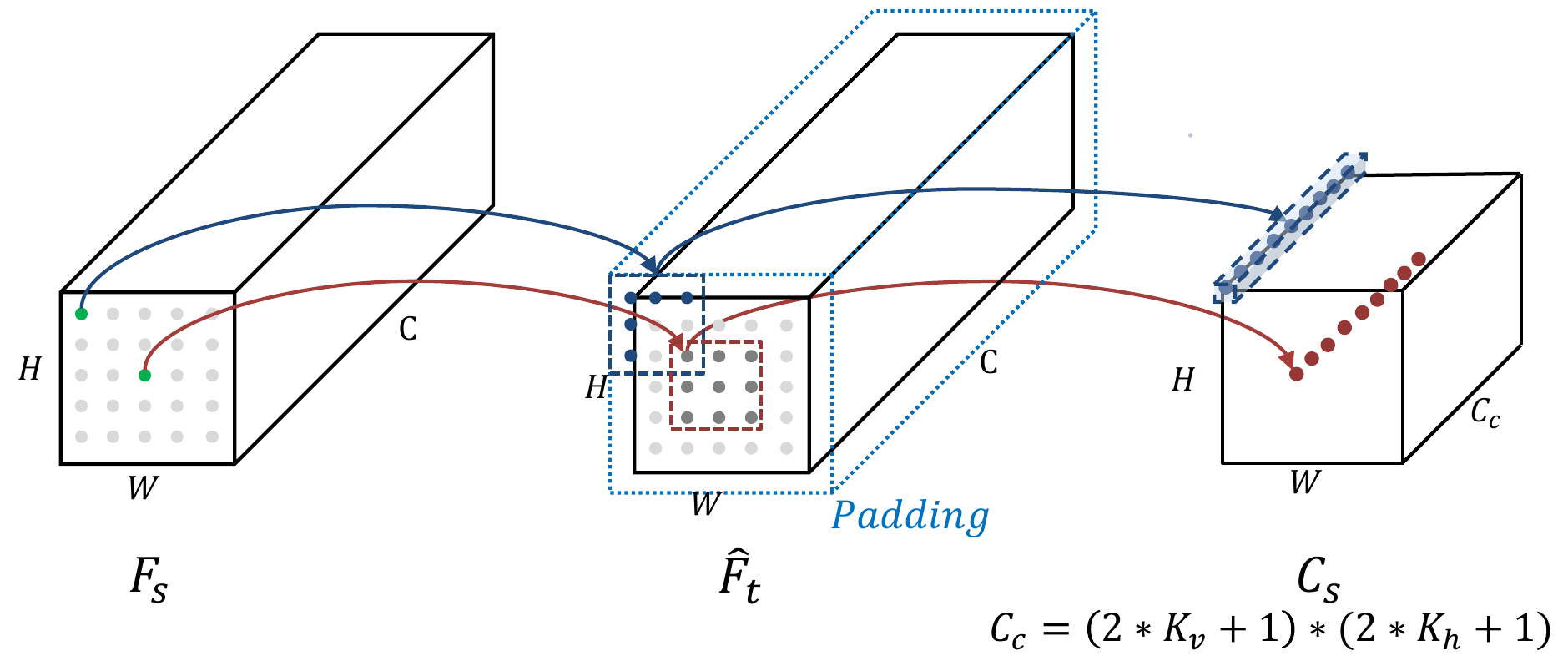}   
\caption{Correlation and marginalization. The marginalization process creates a 3D correlation map ${C_{s}}$ for feature from ${F_{s}}$ to ${F_{t}}$.}
\label{fig:TCD_pixelcorrelation}
\end{figure}

\begin{figure}[t]
\centering
\includegraphics[width=0.95\columnwidth]{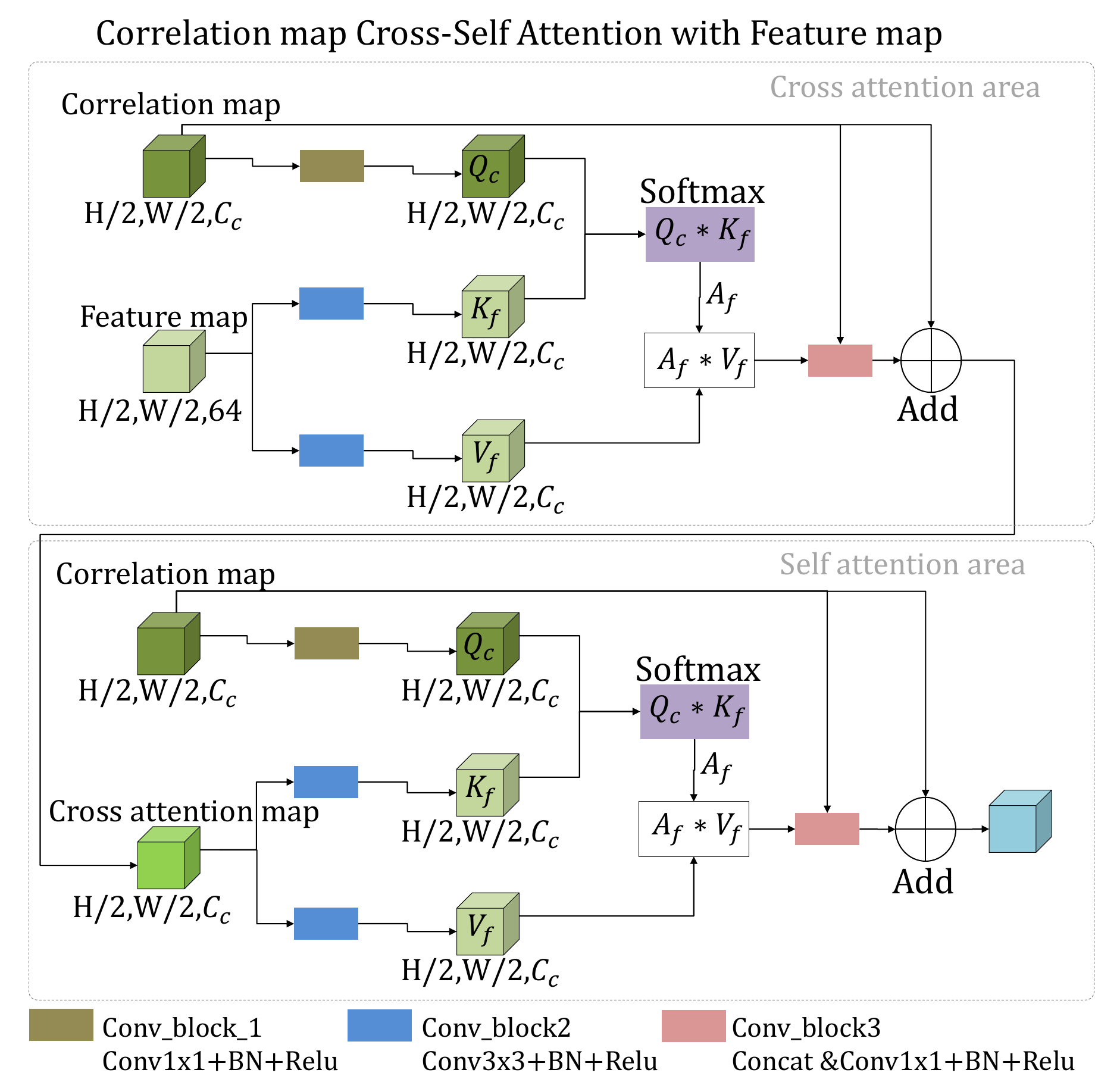}
\caption{Correlation feature map attention process. It shows attention process performed between upsampled correlation maps ${\hat{C}_{s}^{2}}$, ${\hat{C}_{s}^{3}}$, ${\hat{C}_{t}^{2}}$, ${\hat{C}_{t}^{3}}$ and corresponding top level feature maps ${F_s^{1}}, {F_t^{1}}$.}
\label{fig:TCD_cacas}
\end{figure}

As illustrated in \cref{fig:TCD_pixelcorrelation}, we compute the neighboring correlation map within a window of size $(2*K_{v}+1)$, $(2*K_{h}+1)$ at each spatial pixel position in the feature map.
Subsequently, we convert feature maps into marginalized 3D correlation tensors with ${C_{c}}$ channels, where $C_{s}$ represents the correlation from ${F_{s}}$ to ${F_{t}}$, and $C_{t}$ represents the correlation from ${F_{t}}$ to ${F_{s}}$.
A detailed description of the algorithm for computing the $C_{c}$ channel marginalized correlation maps using cosine similarity is provided in \cref{alg:corr}.

\begin{algorithm}[tb]
\caption{Correlation and Marginalized Correlation map}
\label{alg:corr}
\begin{algorithmic}
\STATE \textbf{Input}: {Assume ${F}^{l}_s, {F}^{l}_t$ {source and target feature maps at same scale for ${l}$={2,3}}}\\
\STATE Normalize source and target feature maps along the channel dimension \\
\STATE Pad source and target feature maps by ${K_h,K_v}$ in horizontal and vertical direction on both sides results in ${\hat{F}^{l}_{s}},{\hat{F}^{l}_{t}}$  \\
\STATE Initialize an empty correlation map ${C_{s}, C_{t}}$ of size ${C_{c}}$x${H_{l}}$x${W_{l}}$, where $C_{c}=(2*K_{h}+1)*(2*K_{v}+1)$ \\
\STATE Let ${K_w=(2*K_h+1)}$, ${K_t=(2*K_v+1)}$ \\
\STATE Let $i=1$, $j=1$
\WHILE{$i <= K_w$}
\WHILE{$j <= K_t$}
\STATE ${t1=(\hat{F}^{l}_{t}[i:i+H,j:j+W]*F^{l}_s)}$ 
\STATE ${t2=(\hat{F}^{l}_{s}[i:i+H,j:j+W]*F^{l}_t)}$ 
\STATE ${C_{s}[i*K_w+j,:,:]}$=${\Sigma_{c}t1}$
\STATE ${C_{t}[i*K_w+j,:,:]}$=${\Sigma_{c}t2}$
\ENDWHILE
\ENDWHILE
\STATE \textbf{Output}: {Marginalized correlation map $C_{s},C_{t}$ for each multi-scale feature map set $F^{l}_{s},F^{l}_{t}$}\\
\end{algorithmic}
\end{algorithm}

\subsection{Correlation map cross-self attention with feature map}
In addition to the feature maps \cref{multi_scale_feature_map_att}, we also apply cross-self attention mechanisms to the marginalized lower-level correlation maps.
Prior to applying the attention mechanisms, the marginalized correlation maps ${C_{s}^{2}}$, ${C_{s}^{3}}$ and ${C_{t}^{2}}$, ${C_{t}^{3}}$ from the two levels are deconvolved and upsampled to a size of $(H/2, W/2, C_{c})$ using shared deconvolution and upsampling layers.
Subsequently, the upsampled correlation maps ${\hat{C}_{s}^{2}}$, ${\hat{C}_{s}^{3}}$ and ${\hat{C}_{t}^{2}}$, ${\hat{C}_{t}^{3}}$ are each independently subjected to an attention process with the top-level feature maps ${F_s^{1}}, {F_t^{1}}$.
This attention process, which utilizes transformer attention, is illustrated in \cref{fig:TCD_architecture} and \cref{fig:TCD_cacas}.
Specifically, the 64 channel feature map ${F^{1}}$ serves as the key and value for attention.
To match the channel sizes between the $C_{c}$ channel query correlation map and the encoded feature map ${F^{1}}$, the feature map is squeezed using {Conv\_Block2} to generate new key $K_{f}$ and value $V_{f}$, followed by the attention feature using \cref{eq:TCD_attention}. 
Finally, the input correlation map and the queried feature map $A_{f}*V_{f}$ are merged using {Conv\_Block3} and added to the input correlation map.
Next, cross-self attention is applied to the input correlation map using a similar process to cross-attention, where the key and value features are derived from the cross-attention output, as depicted in the lower box of \cref{fig:TCD_cacas}.
This process is employed between the image-encoded feature $F^{1}$ and all marginalized lower-level correlation maps, resulting in a total of four iterations.

\subsection{Segmentation maps}

The final segmentation maps are generated by integrating correlation feature maps from the attention module within each Decoder. 
These maps undergo convolutional upsampling and Sigmoid activation, yielding two-way semantic segmentation maps ${{S_{st}, S_{ts}}}$ from two parallel decoders, representing changes from $I_s$ to $I_t$ and from $I_t$ to $I_s$.

\subsection{Loss function}

Assuming $I_s$, and $I_t$ are the source and target image, respectively, the segmentation loss function combines dice Loss $L_d$ and binary cross-entropy (BCE) loss $L_{bce}$ as follows:

\begin{equation}
\ L_{st}(I_s,I_t) = W_d*L_d(S_{st},G_{st})+L_{bce}(S_{st},G_{st}) 
\end{equation}
$S_{st}$, $G_{st}$, and $W_{d}$ denote the predicted segmentation map, ground truth segmentation map, and scale factor, respectively.
The dice Loss $L_d$ and BCE loss $L_{bce}$ are formulated as follows:
\begin{equation}
\ L_{d}={1-\frac{2*y_i*p_i+\tau}{y_i+p_i+\tau}} ,\quad \tau = 1
\end{equation}
\begin{equation}
{L_{bce}=-{(y_i\log(p_i) + (1 - y_i)\log(1 - p_i))}}
\end{equation}
For each pixel position $i$, $y_i$ and $p_i$ denote the corresponding ground truth and predicted values, respectively.
Our model generates two segmentation maps; hence, the total segmentation loss is computed by averaging the losses associated with each map. 
The overall loss function is subsequently defined as follows:
\begin{equation}
\ {L_{seg}={0.5*L_{st}(I_s,I_t)+ 0.5*L_{ts}(I_t,I_s) }}
\end{equation}

\subsection{Training with synthetic data}

\begin{figure}[t]
\centering  
\includegraphics[width=0.85\columnwidth]{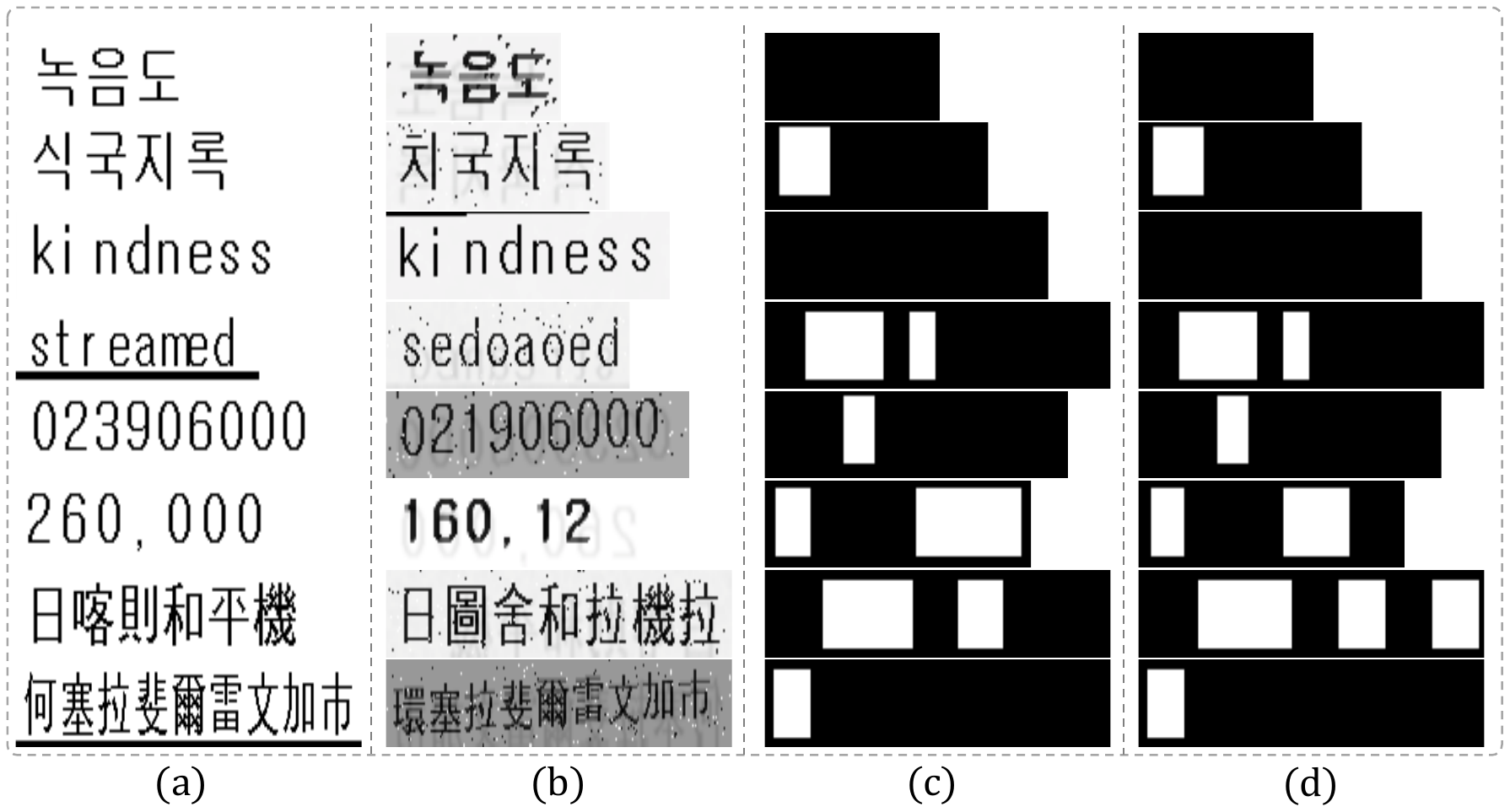}   
\caption{Sample of synthetic training data : (a) source, (b) target, (c) segmentation ground truth from source to target, (d) segmentation ground truth from target to source.}
\label{fig:TCD_trainsample}
\end{figure}

The input to our TCD model comprises pairs of unit text images, where the ground truth contains the changed area information.
To train our model, we generate a dataset using a synthetic image generator based on a diverse text corpus containing English, Korean, Chinese, numerical characters, and special symbols.
Given that our TCD approach is designed as a two-way segmentation task, our ground truth annotations comprise bidirectional segmentation maps.
We simulate the task of comparing contract documents, which often involves comparing machine-readable documents with scanned documents, in our data-generation process.
Specifically, we generate the source image using a fixed background, whereas the target image is generated with a random background to mimic the variability in the image quality of scanned documents.
Examples of the samples generated by our data generator are presented in \cref{fig:TCD_trainsample}.
The sample data is generated as a pseudo-segmentation map, where the value of a rectangular region corresponding to the maximum width and height of a changed character is set to 1, and the remaining areas are set to 0. 
The data generator is designed to produce images with fixed heights but varying widths, thus supporting dynamic unit sizes rather than relying on a fixed size 
During batch training, we create uniformly sized data by padding background pixels to the right side of text images. 
Both source and target images are generated, applying various real-world document changes encountered during scanning, including bleed-through, blurring, noise, scaling, underlining, overlining, crossing, and rotation.
To ensure balance within each batch of data, every batch consists of equal numbers of unchanged and changed data pairs. 
The unchanged pairs are generated using identical text corpora, whereas changed pairs are created by randomly modifying certain characters at arbitrary positions.

\section{Experiments}
\label{Experiments}

For benchmarking, we construct a multilingual word-level (or unit-level) image-pair dataset. 
Our benchmark results encompass language-specific comparisons and an aggregate analysis of the entire evaluation dataset. 
Furthermore, we conduct ablation analysis to empirically demonstrate the utility of each module in our proposed framework.

\subsection{Datasets}

There is a lack of public datasets specifically designed for TCD using image comparison.
To address this gap, we create a novel dataset tailored to change detection for text images.

\textbf{Training and validation datasets:} 
We construct a corpus dataset comprising 10,000 words, encompassing English, Korean, Chinese, numbers, and special characters. 
The training data source images are generated from a text corpus, with modified characters randomly selected from other corpus texts.
Each dataset instance consists of a pair of source images and corresponding target change map images.
To ensure balanced data, identical and different image pairs are generated in equal proportions within each batch.
Furthermore, an additional 5,000 text corpus data points are utilized to create a validation dataset.
\begin{figure*}[t]
\centering
\includegraphics[width=1.85\columnwidth]{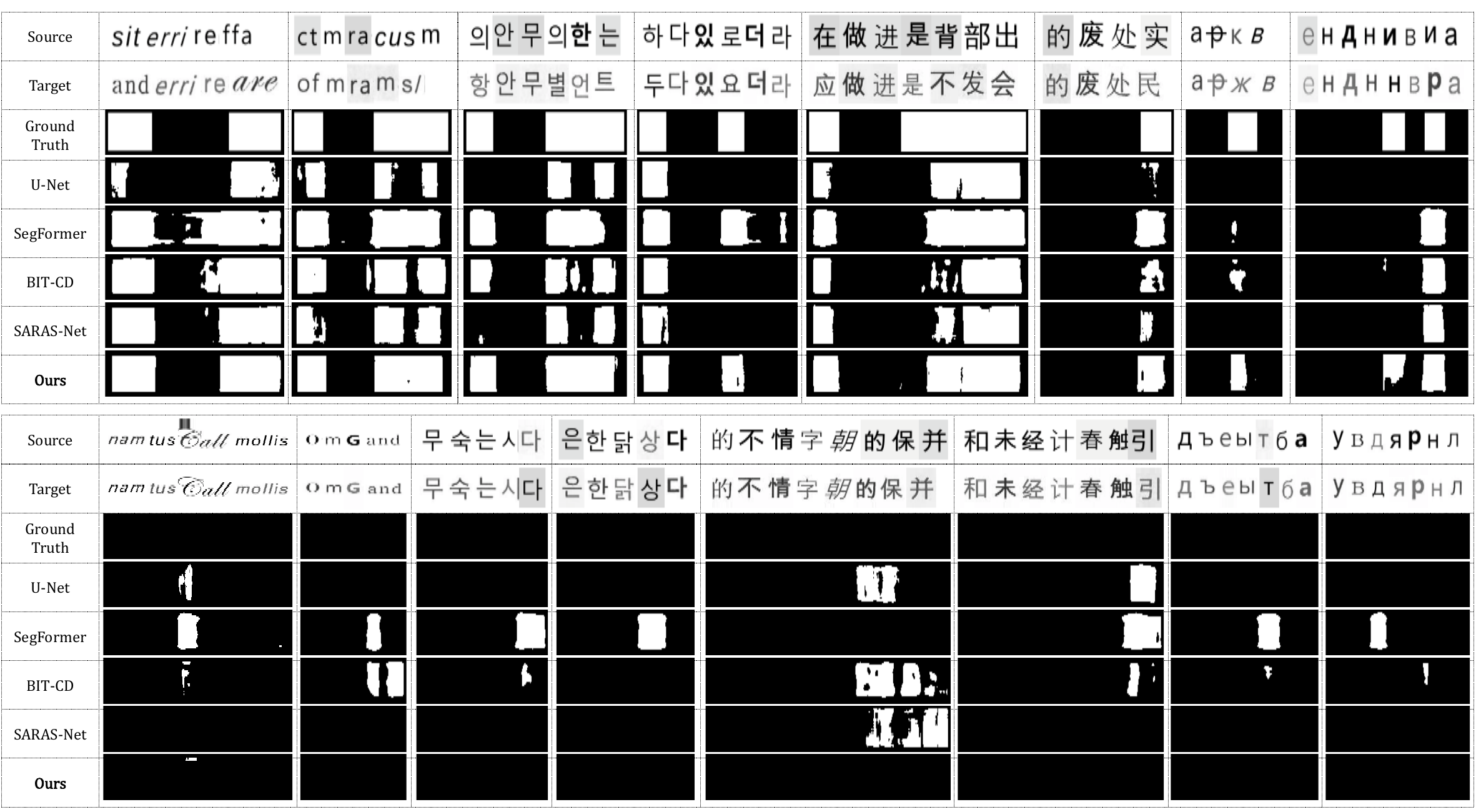}   
\caption{Qualitative results of different segmentation models.}
\label{fig:seg_result}
\end{figure*}

\textbf{Test datasets:}
\label{sec:test_dataset}
We construct two different datasets to serve as benchmarks for evaluating semantic segmentation, and OCR methods. 
Our segmentation test dataset is created by combining actually printed and scanned word pairs extracted from documents written in four languages: English, Korean, Russian, and Chinese.
Using the DUET unit detector \cite{jung2021duet}, we crop the units at the character level.
Subsequently, these cropped units are randomly merged to form concatenated text images, and the synthetic ground truth for segmentation changes is generated in a similar manner to the training dataset.
Using the position in the full image and text ground truth, we produce identical and different pairs.
The resulting dataset consists of 80,000 pairs, with 10,000 identical pairs and 10,000 different pairs for each of the four languages.

Our OCR test dataset is constructed from the public datasets Distorted Document Images (DDI-100)\cite{zharikov2020ddi} and the LRDE Document Binarization Dataset (LRDE DBD)\cite{lazzara2011scribo} and from our own data generated by printing and scanning documents. 
DDI-100 contains English, Russian, and numerical digits, while LRDE features French text. 
In addition, we incorporate Korean, Arabic, Hindi, and Chinese into our dataset.
Paired images with identical content from the original and distorted or scanned versions are used to create the dataset, accompanied by the OCR ground truth. 
Text units are cropped from each document image using ground truth coordinates.
The dataset is then organized according to OCR ground truth, and identical and different pairs are created based on the text. 
Different paired data are randomly selected, ensuring a maximum of four character changes and a maximum length difference of 1. 
A summary of the OCR dataset is presented in \cref{tab:ocr_dataset}.

\begin{table}[t]
\centering
\resizebox{.85\columnwidth}{!}{
\begin{tabular}{l|c|c|c|c|c|c|c|c}
     \toprule
    language    & English & Russian & French & Korean & Chinese & Digits & Arabic & Hindi\\
    \midrule
    Identical        & 50002 & 47462 & 19234 & 10228 & 17538 & 11167 & 3250 & 4457 \\
    Different        & 49998 & 52538 & 30766 & 39772 & 32462 & 38833 & 3250 & 4457 \\
    \midrule
    Total       & 100000 & 100000 & 50000 & 50000 & 50000 & 50000 & 6500 & 8914 \\
    \bottomrule
\end{tabular}
}
\caption{OCR dataset taxonomy}
\label{tab:ocr_dataset}
\end{table}

\subsection{Training configuration}

In our experiment, we fix the height of the training data to 32.
For correlation marginalization, we set the neighboring values $K_{v}, K_{h}$ to 2 and 4 for the $H/4$ features, and 1 and 2 for the $H/8$ features, respectively.
We also scale the segmentation loss by a scale factor $W_d$ of 10 during training.
Our model is trained from scratch for 200 epochs with a batch size of 8.

\begin{table*}[t]
\centering
\resizebox{1.9\columnwidth}{!}{
\begin{tabular}{l|c c c c c| c c c c c | c c c c c | c c c c c | c c c c c}
    \toprule
    language &  \multicolumn{5}{c|}{English} & \multicolumn{5}{c|}{Korean} & \multicolumn{5}{c|}{Chinese} & \multicolumn{5}{c|}{Russian} & \multicolumn{5}{c}{Avg.}\\
    \midrule
    Model &  Pre. & Rec.    & F1  & IoU      &  OA  &  Pre. & Rec.    & F1 & IoU      &  OA  &  Pre. & Rec.    & F1  & IoU      &  OA  &  Pre. & Rec.    & F1 & IoU      &  OA &  Pre. & Rec.    & F1  & IoU      &  OA  \\
    \midrule
        U-Net\cite{ronneberger2015u}        & \textbf{65.8} & 54.5 & 59.6 & 42.5 & 83.2     & 63.1 & 71.8 & 67.2 & 50.6 & 85.6       & \textbf{63.5} & 67.8 & 65.5 & 48.7 & 84.5       & 63.6 & 68.7 & 66.0 & 49.3 & 85.7                                             & \textbf{64.0} & 65.7 & 64.6 & 47.8 & 84.8 \\
        SegFormer-B5\cite{xie2021segformer} & 61.2 & \textbf{82.7} & 70.3 & 54.2 & 84.1     & 61.1 & 83.5 & 70.6 & 54.5 & 85.7       & 62.1 & \textbf{78.0} & 69.2 & 52.9 & 84.9       & 59.1 & 79.7 & 67.9 & 51.4 & 84.8                                    & 60.9 & 81.0 & 69.5 & 53.2 & 84.9 \\
        BIT-CD\cite{chen2021remote}       & 62.8 & 77.5 & 69.4 & 53.1 & 84.4     & \textbf{64.4} & 73.9 & 68.8 & 52.4 & 86.3       & 62.9 & 61.1 & 62.0 & 44.9 & 83.8       & \textbf{65.6} & 75.8 & 70.3 & 54.3 & 87.1                                    & 63.9 & 72.1 & 67.6 & 51.2 & 85.4 \\
        SARAS-Net\cite{chen2023saras}    & 62.5 & 79.7 & 70.0 & 53.9 & 84.5     & 62.3 & 70.4 & 66.1 & 49.3 & 85.2       & 62.0 & 65.9 & 63.9 & 46.9 & 83.8       & 61.8 & 66.0 & 63.8 & 46.8 & 84.9                                                      & 62.1 & 70.5 & 65.9 & 49.2 & 84.6 \\
        TCD(ours)    & 64.1 & 81.3 & \textbf{71.7} & \textbf{55.9} & \textbf{85.4}     & 63.2 & \textbf{84.2} & \textbf{72.2} & \textbf{56.5} & \textbf{86.7}       & 63.3 & 76.5 & \textbf{69.3} & \textbf{53.0} & \textbf{85.3}       & 64.2 & \textbf{83.6} & \textbf{72.6} & 57.0 & \textbf{87.3}                           & 63.7 & \textbf{81.4} & \textbf{71.5} & \textbf{55.6} & \textbf{86.3} \\
    \bottomrule
\end{tabular}
}
\caption{Quantitative results of segmentation benchmark. Average results show that TCD achieves the highest performance in terms of F1 and Acc by 2\% and 1\% from the SotA, respectively.}
\label{tab:seg_result}
\end{table*}

\begin{table*}
\centering
\resizebox{1.9\columnwidth}{!}{
\begin{tabular}{l|c c c c|c c c c | c c c c | c c c c | c c c c | c c c c | c c c c | c c c c | c c c c}
    \toprule
    language &  \multicolumn{4}{c|}{English} & \multicolumn{4}{c|}{Korean} & \multicolumn{4}{c|}{Chinese} & \multicolumn{4}{c|}{Russian} & \multicolumn{4}{c|}{French} & \multicolumn{4}{c|}{Hindi}  & \multicolumn{4}{c|}{Arabic} & \multicolumn{4}{c|}{Digits} & \multicolumn{4}{c}{Avg.}\\
    \midrule
     Method &  Pre. & Rec.  & F1 &  Acc  &   Pre. & Rec.    & F1 &  Acc. &  Pre. & Rec.  & F1 &  Acc   &  Pre. & Rec.  & F1 &  Acc   &  Pre. & Rec.  & F1 &  Acc  &  Pre. & Rec.  & F1 &  Acc &  Pre. & Rec.  & F1 &  Acc  &  Pre. & Rec.  & F1 &  Acc   &  Pre. & Rec.  & F1 &  Acc   \\
    \midrule
        Tesseract OCR\cite{smith2007overview}                       & 87.7 & 99.9 & 93.4 & 92.9                 & 97.8 & \textbf{100.0} & 98.9 & 98.2                   & 84.2 & \textbf{100.0} & 91.4 & 89.7                   & 89.7 & \textbf{99.8} & 94.5 & 93.9                    & 94.2 & \textbf{99.7} & 96.8 & 96.0                                                                                                                        & 99.6 & 51.2 & 67.6 & 52.4                 & 99.8 & 53.5 & 69.6 & 56.5                 & 87.3 & 99.9 & 93.2 & 88.7                                                         & 92.5 &  88.8 & 88.2 & 83.5 \\
        Multiplex OCR\cite{huang2021multiplexed}                    & 56.2 & 98.3 & 71.5 & 60.9                 & 84.6 & 95.7 & 89.8 & 82.7                             & 75.4 & 87.2 & 80.9 & 73.2                             & 55.6 & 98.0 & 70.9 & 57.8                             & 65.5 & 98.5 & 78.7 & 67.1                                                                                                                                 & 43.2 & 59.9 & 50.2 & 57.1                 & 65.3 & 58.5 & 61.7 & 59.5                 & 87.1 & 98.3 & 92.3 & 87.3                                                         & 66.6 &  86.8 & 75.6 & 68.2 \\
        PPOcr v3\cite{li2022pp}                                     & 99.3 & \textbf{100.0} & \textbf{99.6} & \textbf{99.6}     & 99.5 & 99.9 & \textbf{99.7} & \textbf{99.5}           & 99.0 & \textbf{100.0} & 99.5 & 99.3                   & 79.8 & 99.4 & 88.5 & 86.5                             & 96.4 & 99.6 & 98.0 & 97.5                                                                                                                 & 99.8 & 66.9 & \textbf{80.1} & \textbf{75.4}                 & 99.9 & 57.7 & 73.1 & 63.3                  & \textbf{99.8} & \textbf{100.0} & \textbf{99.9} & \textbf{99.9}                    & \textbf{96.7} &  90.4 & 92.3 & 90.1 \\
        TrOCR${_{BASE}}$\cite{li2023trocr}                          & 91.5 & 97.4 & 94.4 & 94.2                                 & 83.5 & 99.6 & 90.9 & 84.1                             & 69.3 & 99.7 & 81.6 & 70.8                             & 60.0 & 99.6 & 74.9 & 64.9                             & 86.4 & 98.0 & 91.8 & 89.3                                                                                                                 & \textbf{99.9} & 53.9 & 70.0 & 57.3                 & \textbf{100.0} & 52.6 & 68.9 & 54.9         & 92.8 & 99.9 & 96.3 & 93.9                                                         & 85.4 & 87.6 & 83.7 & 76.2  \\
        TCD(ours)                                                   & \textbf{99.4} & 99.5 & 99.4 & 99.4                        & \textbf{100.0} & 99.4 & \textbf{99.7} & \textbf{99.5} & \textbf{99.7} & 99.9 & \textbf{99.8} & \textbf{99.7}  & \textbf{99.4} & 99.4 & \textbf{99.4} & \textbf{99.4}  & \textbf{98.9} & 97.4 & \textbf{98.1} & \textbf{97.7}                                                                                      & 63.2 & \textbf{99.3} & 77.2 & 70.7      & 72.1 & \textbf{99.6} & \textbf{83.7} & \textbf{80.5}         & 99.7 & 99.8 & 99.7 & 99.6                   & 87.5 &  \textbf{99.3} & \textbf{94.6} & \textbf{93.3}  \\
    \bottomrule    
\end{tabular}
}
\caption{Quantitative results of OCR benchmark. Benchmark results show that TCD achieves little difference across various languages.}

\label{tab:ocr_result}
\end{table*}

\subsection{Evaluation}

To use the segmentation test dataset as a benchmark, we adopt a range of performance metrics commonly used in semantic segmentation and change detection models, including precision, recall, the F1 score, IoU, and overall accuracy. 
Notably, our evaluation dataset consists of both identical- and different-pair data because these metrics assess performance at the pixel level across the entire dataset.
To evaluate the OCR test data, we employ binary classification performance scores for both the identical and different pairs. 
Specifically, we calculate precision, recall, the F1 score, and accuracy for all OCR benchmarking models using a classification confusion matrix. 

\subsection{Benchmark results}

To evaluate the performance capabilities of our model, we select several well-known and SotA semantic segmentation and change detection models, including U-Net \cite{ronneberger2015u}, SegFormer \cite{xie2021segformer}, BIT-CD \cite{chen2021remote}, and SARAS-Net \cite{chen2023saras}. 
Additionally, we include the OCR methods Tesseract OCR \cite{smith2007overview}, Multiplex OCR \cite{huang2021multiplexed}, PPOcr v3 \cite{li2022pp} and TrOCR \cite{li2023trocr}.
Typically, semantic segmentation and change detection models are not trained on text datasets.
To ensure a fair comparison, we retrain these models using our synthetic training data generator in a manner similar to our model for 200 epochs each with the default settings.
To evaluate the OCR models, we use publicly available pre-trained models for each language.

We define true-positive (TP) as instances where changes are correctly predicted as different across all experiments, since our proposed model aims to identify changed pairs. 
Recall measures the ratio of correctly predicted as different among actual changed data, reflecting the model's ability to detect true changed pairs in this experimental setup.
Moreover, precision measures how often the model incorrectly identifies identical pairs as different.

\Cref{tab:seg_result} presents a performance comparison between our model and the segmentation and change detection methods across the four language segmentation datasets.
Our model consistently outperforms all SotA methods, achieving an average improvement of 2 points for the F1 score and IoU, and 1 point for accuracy.
Moreover, our model maintains a similar performance across all four language datasets.
\Cref{fig:seg_result} visualizes the prediction results for the compared methods using these datasets.
The top rows of \cref{fig:seg_result} depict scenarios with changed text, while the bottom rows show scenarios with unchanged text. 
White areas indicate changed text, and black areas indicate unchanged text.
Although our model generates two segmentation maps, only the maximum result of the two outputs is shown for comparison with the benchmark single-output models.
The actual source and target input pairs are not the same size. 
Therefore, prior to inference, the height of each input is resized, and the width is right-padded to ensure uniform dimensions. 
From the prediction results, it can be observed that our model generates segmentation maps that are sharper, less noisy, and closer to the ground truth for various examples of text changes and text change positions.
In contrast, other methods often fail to detect real changes in some cases.
For identical text image input pairs, all other methods produce noisy and false-positive change maps.

Similarly, \cref{tab:ocr_result} presents a performance comparison with OCR methods across seven languages and numerical digit datasets. 
We classify paired text images as either identical or different. 
An input text-image pair is classified as identical if the number of changed pixels is zero; otherwise, it is classified as different.
Typically, OCR results tend to depend on the language.
Most methods perform well on digits and English, but their performance is slightly worse for some languages, such as Russian, Hindi, and Arabic.
While most OCR methods exhibit good results, each language requires its own model weights.
Notably, since TrOCR is primarily trained on English corpus, it requires retraining for application to other languages.
Although not discussed in detail here, OCR recognition errors can also occur, leading to high recall for the detection of differences but relatively low precision in correctly identifying identical pairs.
Multiplex OCR uses a single weight, but performance varies significantly depending on the language.
In contrast, our model demonstrates high performance regardless of the language.

\subsection{Ablation analysis}

We conduct an ablation study on various modules of our TCD model using our segmentation benchmark dataset.
The modules analyzed include correlation and marginalization (CM), encoder feature map attention (FA), decoder correlation map attention (CA), and one-way (OW) versus two-way (TW) segmentation maps. 
Additionally, we compare the performance of a baseline model created by removing all specified modules and replacing the CM module with convolutional layers. 
This allows us to evaluate the impact of each module by progressively adding or removing them from the model.
All models are trained for 200 epochs each and then evaluated on the segmentation dataset. 

\begin{table}[t]
\centering
\resizebox{0.75\columnwidth}{!}{
\begin{tabular}{l|c c c c | c c c c c}
    \toprule
    model/module&CM & FA & CA & TW    & Pre. & Rec. & F1  & IoU & OA \\
    \midrule
    TCD v1& X  & X & X &   \checkmark & 62.8 & 70.4 & 66.3 & 49.7 & 84.8 \\
    TCD v2&\checkmark  & X & X &  \checkmark  & 62.2 & 80.3 & 70.1 & 54.0 & 85.4 \\
    TCD v3&\checkmark &  \checkmark & X &  \checkmark  & 62.7 & 76.9 & 69.1 & 52.8 & 85.4 \\
    TCD v4&\checkmark & X &  \checkmark & \checkmark & \textbf{65.1} & 75.7 & 69.9 & 53.8 & \textbf{86.3} \\
    TCD v5&\checkmark &  \checkmark &  \checkmark & X  & 64.9 & 76.9 & 70.3 & 54.3 & \textbf{86.3} \\
    ours  &\checkmark & \checkmark & \checkmark & \checkmark  & 63.7 & \textbf{81.4} & \textbf{71.5} & \textbf{55.6} & \textbf{86.3} \\
    \bottomrule
\end{tabular}
 }
\caption{Ablation study results to show the effect of each module (all language average) on segmentation dataset.} 
\label{tab:able}
\end{table}

\textbf{Correlation and marginalization (CM):} To investigate the effect of the CM module, we first train TCD v1 by replacing the CM module with basic Conv layers.
As shown in \cref{tab:able}, the addition of the CM module (v2) results in improvements in recall, the F1 score, IoU, and OA by 10\%, 4\%, 4\%, and 1\%, respectively.
This shows that the correlation map of features can extract more meaningful information for change detection compared to a simple feature map.

\textbf{Cross-self attention (FA or CA):} FA is applied across the source and target to enhance the feature map, improving features useful for change detection and scale variations between two images.
CA modules are used on the lower-resolution correlation map to improve the change segmentation map.
The TCD v3 and TCD v4 rows indicate that these modules affect precision and overall accuracy, despite a degradation in recall.

\textbf{Two-way segmentation (TW):} We train TCD v5 using a one-way segmentation map, similar to other semantic segmentation models, by taking the maximum output from the two-way model and training it with the single segmentation map loss instead of the TCD two-way loss.
When comparing the performance of TCD v5 and our model, our model’s precision decreased slightly, but the other metrics improved.
Our model’s two-way segmentation result is useful for analyzing not only simple changes but also the nature of these changes.

\section{Conclusion}
In this paper, we introduce a TCD model that enables the comparison of image documents regardless of the language. 
Our model offers language independence by focusing on image comparison rather than text recognition. 
The TCD architecture incorporates multi-scale features and correlation marginalized maps to detect text image changes at a granular unit level within documents. 
Notably, our model operates robustly with various alterations to the text unit images without the need for preprocessing steps such as text or scale alignment. 
Moreover, we incorporate a correlation map integrated with feature map cross-self transformer-based attention mechanisms to enhance the change segmentation accuracy.
The experimental results illustrate the robustness and generalizability of our model across multilingual documents, including those not represented in the training corpus. 
Benchmarking demonstrates that our approach consistently outperforms SotA semantic segmentation models by a significant margin. 
Moreover, our model achieves comparable performance to OCR methods tailored for individual languages.
\clearpage

{\small
\bibliographystyle{ieee_fullname}
\bibliography{egbib}
}

\clearpage

\onecolumn
\begin{appendices}
\section{Document change detection application whole process using TCD}
\label{Appendix}

In the supplementary material, we describe how our proposed method is applied to justify the necessity of our proposed model.
We develop a contract document change detection (DCD) application using our proposed model, which is currently being used by the legal review team.
Our application detects image pairs for comparison through several steps.
The overall DCD process is as follows:

\subsection{Preprocess}
 
\begin{figure}[ht!]
\centering
\includegraphics[width=0.9\columnwidth]{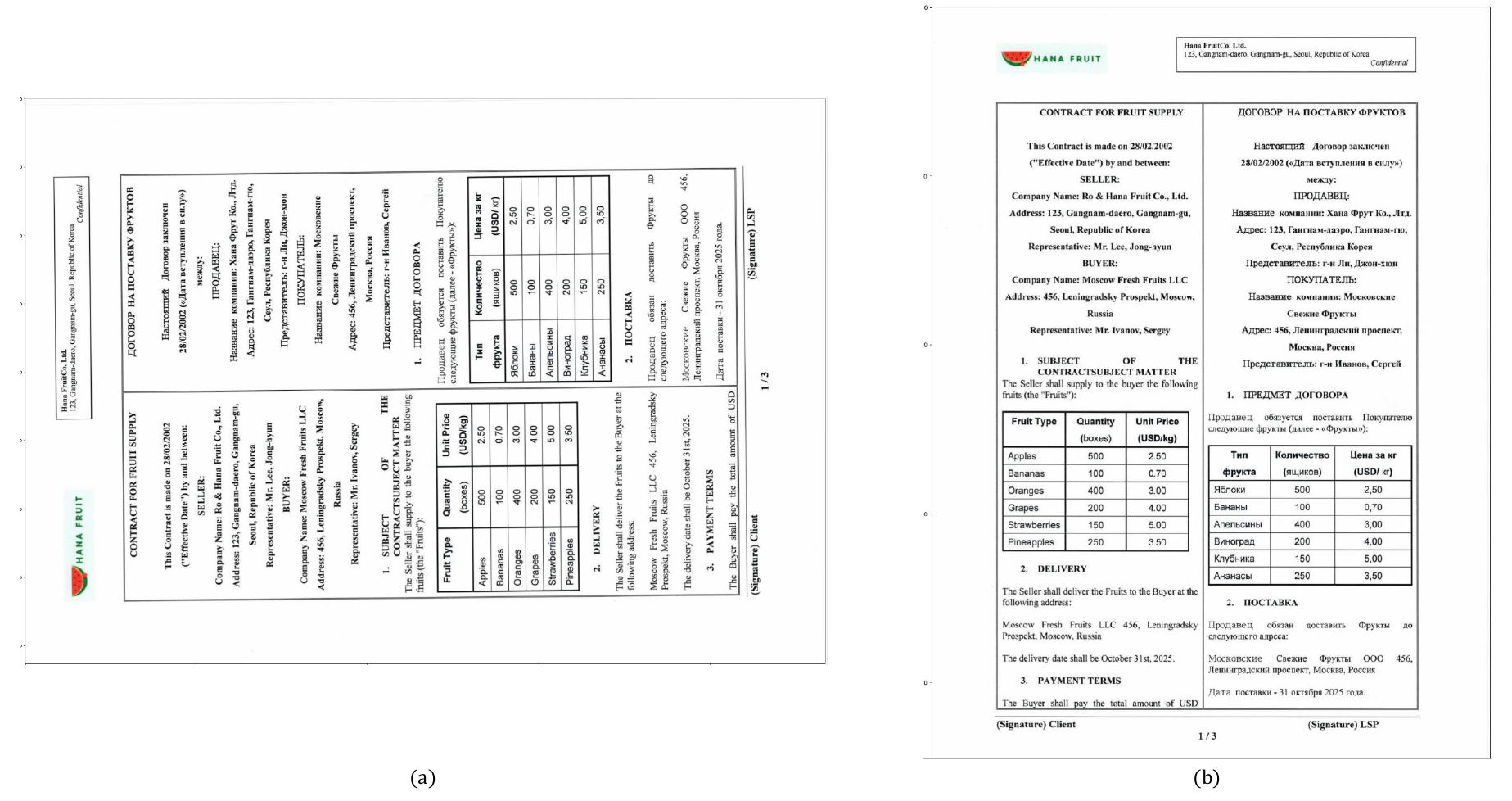}   
\caption{Pre-processing result of a scanned document}
\label{fig:preprocess}
\end{figure}

Typically, we receive the original contract document and its scanned version as inputs, where the scanned version may be rotated or skewed upon entry as illustrated in \cref{fig:preprocess} $(a)$.
In addition, the aspect ratio may vary depending on the device settings. 
To address this issue, we adjusted the scanned version to align with the orientation and scale of the original document as depicted in \cref{fig:preprocess} $(b)$.

\subsection{Layout detection}

\begin{figure}[ht!]
\centering
\includegraphics[width=0.9\columnwidth]{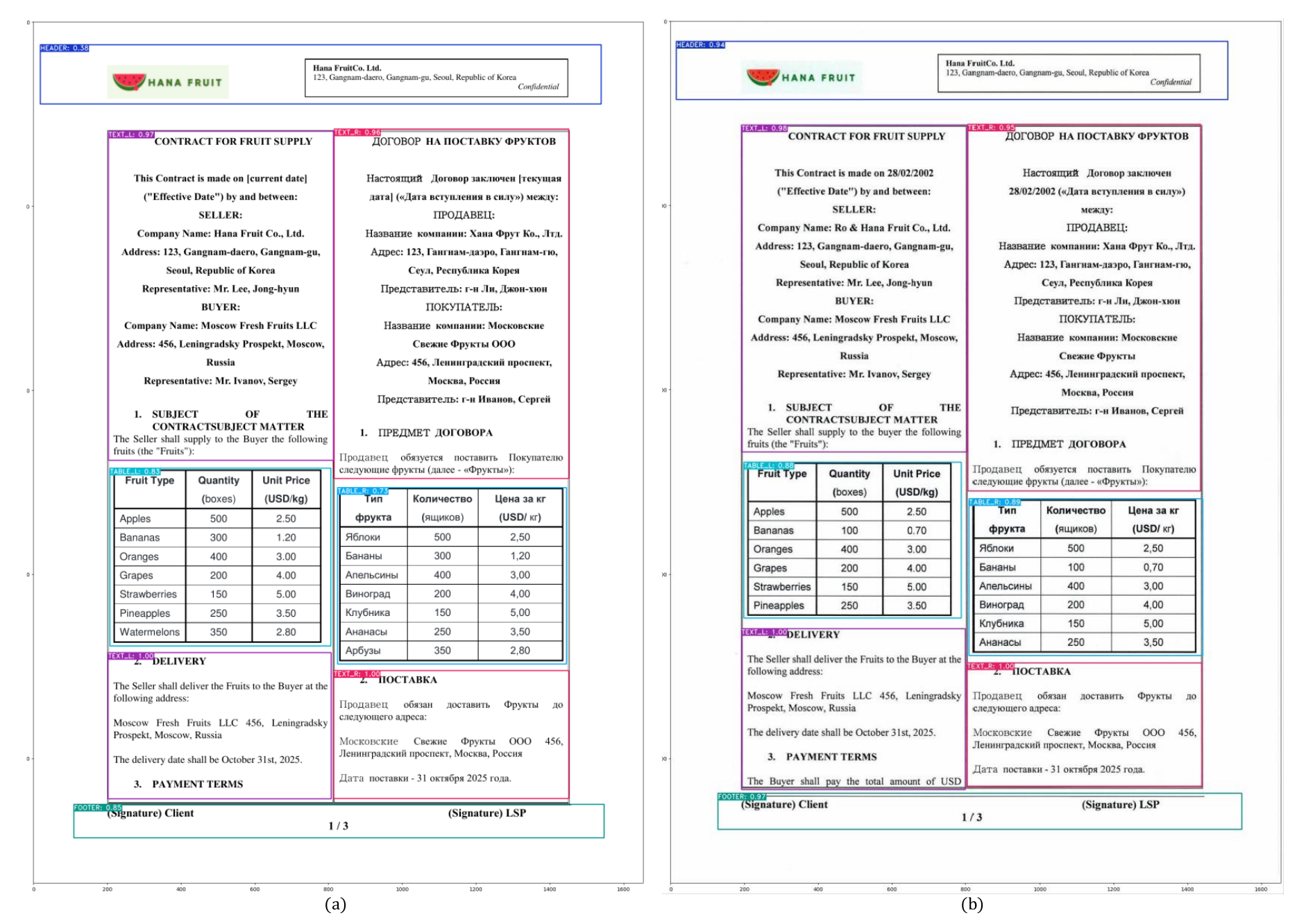}   
\caption{Layout detection results of original and adjusted scanned documents}
\label{fig:layout_detection}
\end{figure}

Contract documents often involve bilingual content, typically presented in a two-column layout. 
Since the documents contain different languages and layouts, we separate the layout to facilitate comparison.  
In our application, we classify the layout into distinct elements such as text, tables, headers, and footers, allowing for precise analysis.
Furthermore, text and table layouts are divided into left, right, and center based on their position.
The layout detection results for the original document and its scanned version are shown respectively in \cref{fig:layout_detection} as $(a)$ and $(b)$.

\subsection{Unit detection}

\begin{figure}[ht!]
\centering
\includegraphics[width=0.9\columnwidth]{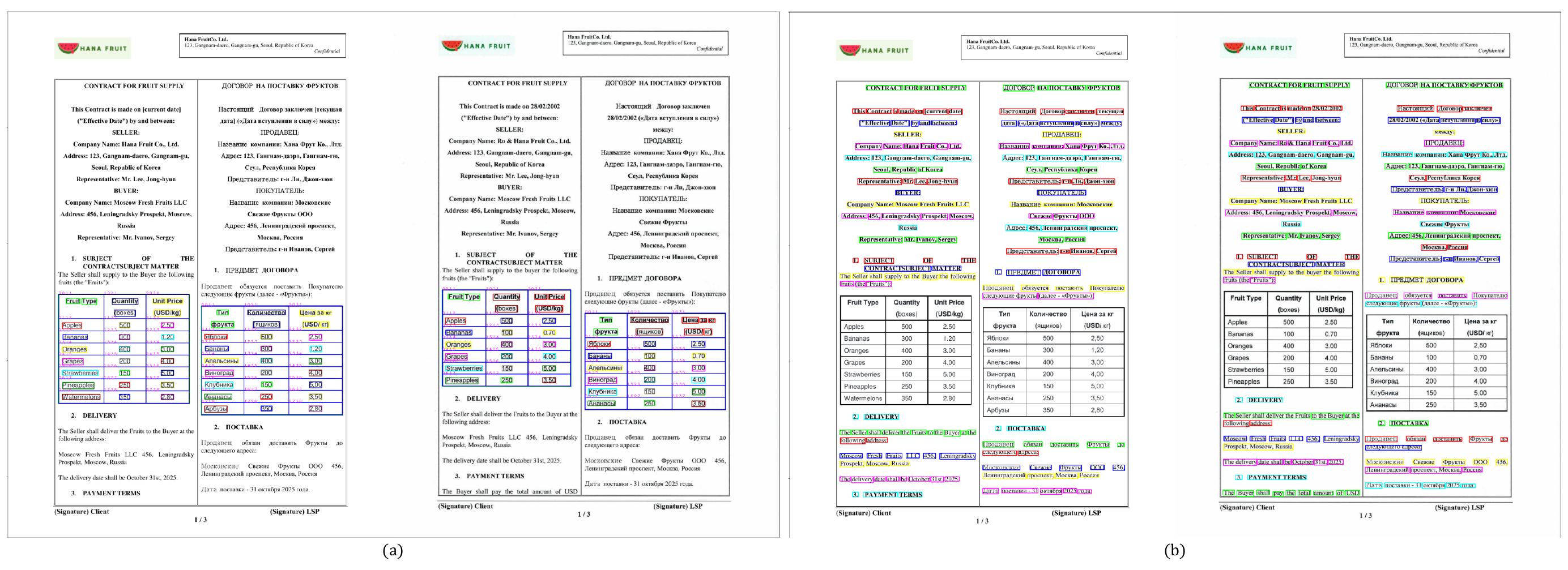}
\caption{Unit detection results}
\label{fig:unit_detection}
\end{figure}

We identify units within each layout element after detecting the layout.
These units are typically detected at the word level, although variations may occur due to factors such as spacing.
The result of this process is shown in \cref{fig:unit_detection}, where $(a)$ and $(b)$ illustrate the results of table and text layout, respectively.
Subsequently, we compare these detected units in pairs, and units that are differently detected in the original and scanned versions undergo a split-and-merge process for accurate comparison. 
Ultimately, this enables us to detect the final changed regions.
The matching process only takes place between units belonging to the same layout element.

\subsection{Two-way segmentation}

\begin{figure}[ht!]
\centering
\includegraphics[width=0.9\columnwidth]{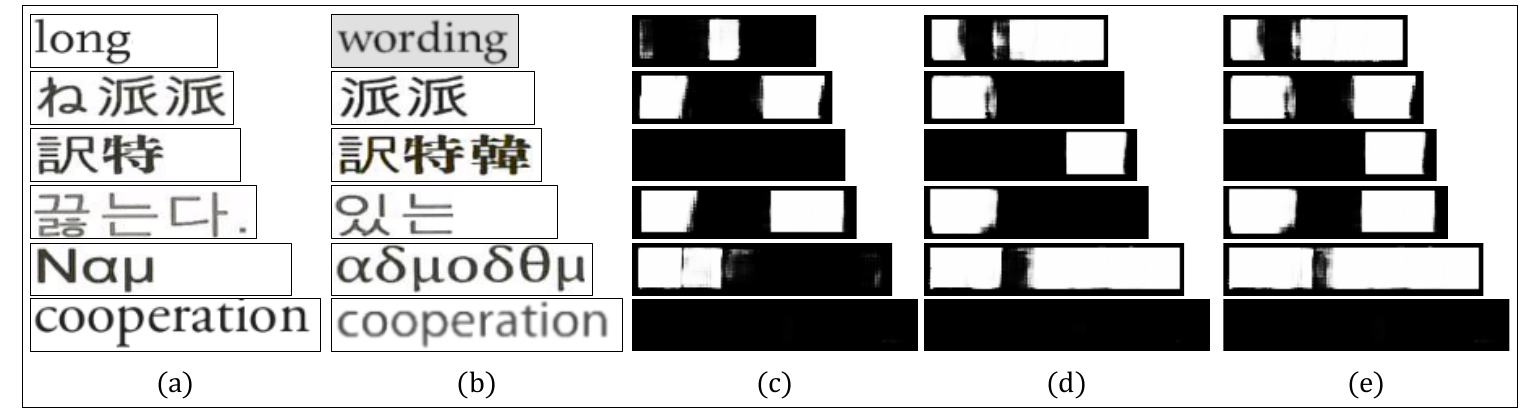}
\caption{Two-way segmentation map : (a) source image, (b) target image, (c) changes from the source perspective, (d) changes from the target perspective, (e) maximum value segmentation map of (c) and (d)}
\label{fig:two_way}
\end{figure}

Our proposed method generates segmentation maps in both directions, \eg, from the source to target and vice versa.
Since the final decision is made through binary classification, we utilize the maximum values of these two segmentation maps.
By analyzing the segmentation maps in both directions, we can identify areas where text has been inserted or deleted.
\Cref{fig:two_way} illustrates the results of the proposed method using the test data samples in \ref{sec:test_dataset}.
Specifically, $(a)$ and $(b)$ represent the source (original) and target (scanned) pairs, respectively, while $(c)$ and $(d)$ denote the changes from the source to target segmentation map and vice versa.
Furthermore, $(e)$ displays the maximum value segmentation map of $(c)$ and $(d)$.
In the third row, for example, since the existing characters do not change in $(a)$, nothing is activated in $(c)$. 
However, from the perspective of $(b)$, the last character is removed, which activates the segmentation map containing the last character area in $(d)$. 
In this case, we infer an insertion from source to target.

\section{Supplementary experimental results}

In Sec.4, we only conduct experiments with text image data, because its objectives diverge from those of traditional change detection approaches.
However, we also conduct additional experiments, and the results are included in the supplementary materials due to page length constraints in the paper.

\subsection{Segmentation results for Hindi and Arabic}
 
\begin{figure}[t]
\centering
\includegraphics[width=0.95\columnwidth]{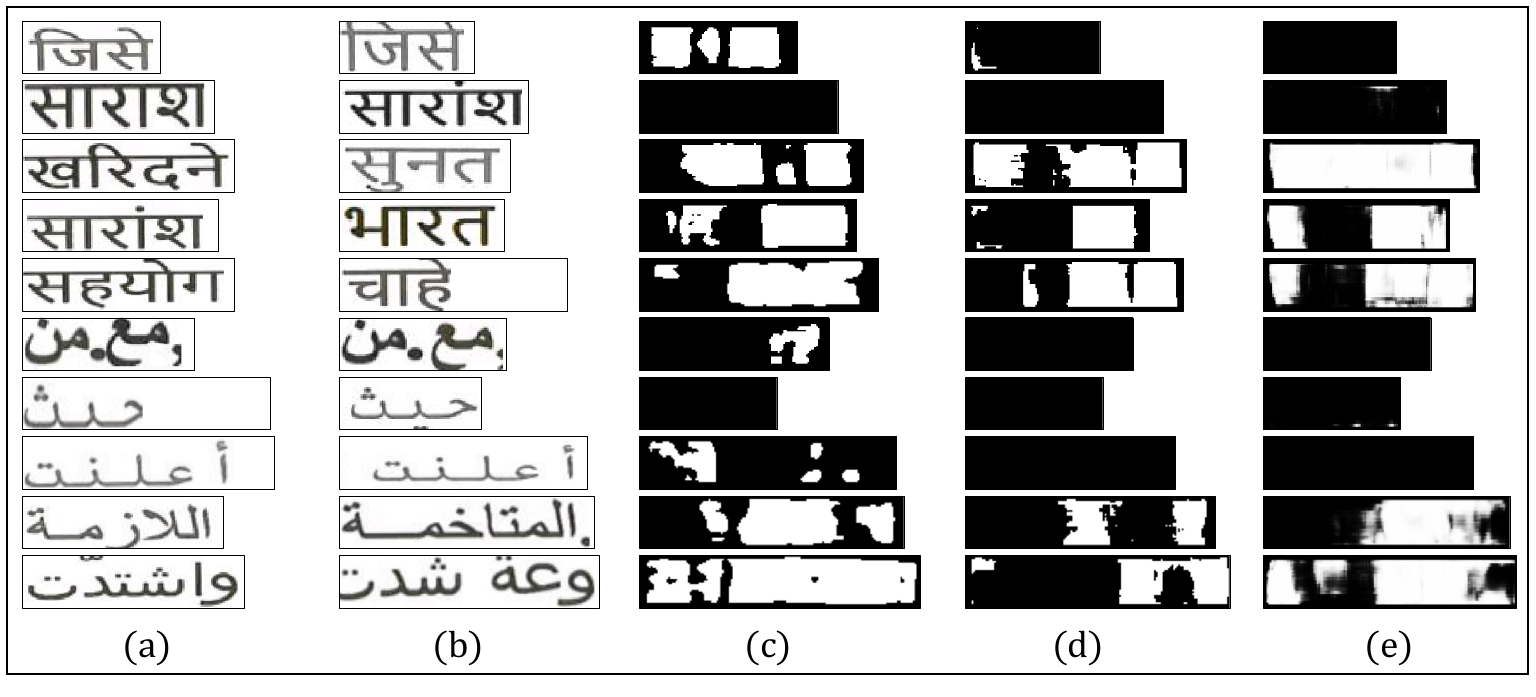}   
\caption{Qualitative comparison of our approach with state-of-the-art methods on Arabic and Hindi dataset: (a) source image, (b) target image, (c) BIT-CD \cite{chen2021remote}, (d) SARAS-net \cite{chen2023saras}, and (e) TCD(ours)}
\label{fig:Arabic_Hindi}
\end{figure}

To compare the performance of our model with SotA change detection models on text images, we generate document images in multiple languages through printing and scanning, and then create word-level image pairs using a unit detector. 
However, the unique linguistic properties of Arabic and Hindi render it exceedingly challenging to produce image pairs using the unit detector.
Due to this limitation, we are unable to create segmentation benchmark datasets for these languages. 
Instead of quantitative evaluation, we utilize OCR test datasets and conduct experiments to qualitatively evaluate the segmentation results for these languages.
The results are shown in the \cref{fig:Arabic_Hindi}.

We conduct experiments comparing the performance of our proposed approach against SotA segmentation models, including BIT-CD \cite{chen2021remote} and SARAS-net \cite{chen2023saras}. 
Notably, our proposed model produces a two-way segmentation map, yielding both source to target and vice versa. 
To facilitate a comprehensive evaluation, we display the maximum values of these two maps in \cref{fig:Arabic_Hindi}.
The BIT-CD model exhibits sensitivity to even slight variations in character ratio and position, often misclassifying such cases as different. 
This limitation is evident in the results shown in \cref{fig:Arabic_Hindi} rows 1, 6, and 8. 
In contrast, SARAS-net demonstrates a converse behavior, struggling to identify discrepancies in different pairs, as observed in \cref{fig:Arabic_Hindi} rows 4, 5, and 9.
Notably, our proposed model demonstrates robustness to character ratio variations and minor truncations, accurately identifying modified regions. 
As evidenced by \cref{fig:Arabic_Hindi}, only our model successfully detects changes in rows 2 and 7. 
Moreover, across most cases, our model excels at identifying modified regions through its two-way analysis, providing the most comprehensive detection of modified regions from both source and target image perspectives.

\subsection{Segmentation results for conventional change detection dataset}
 
\begin{figure}[t]
\centering
\includegraphics[width=0.95\columnwidth]{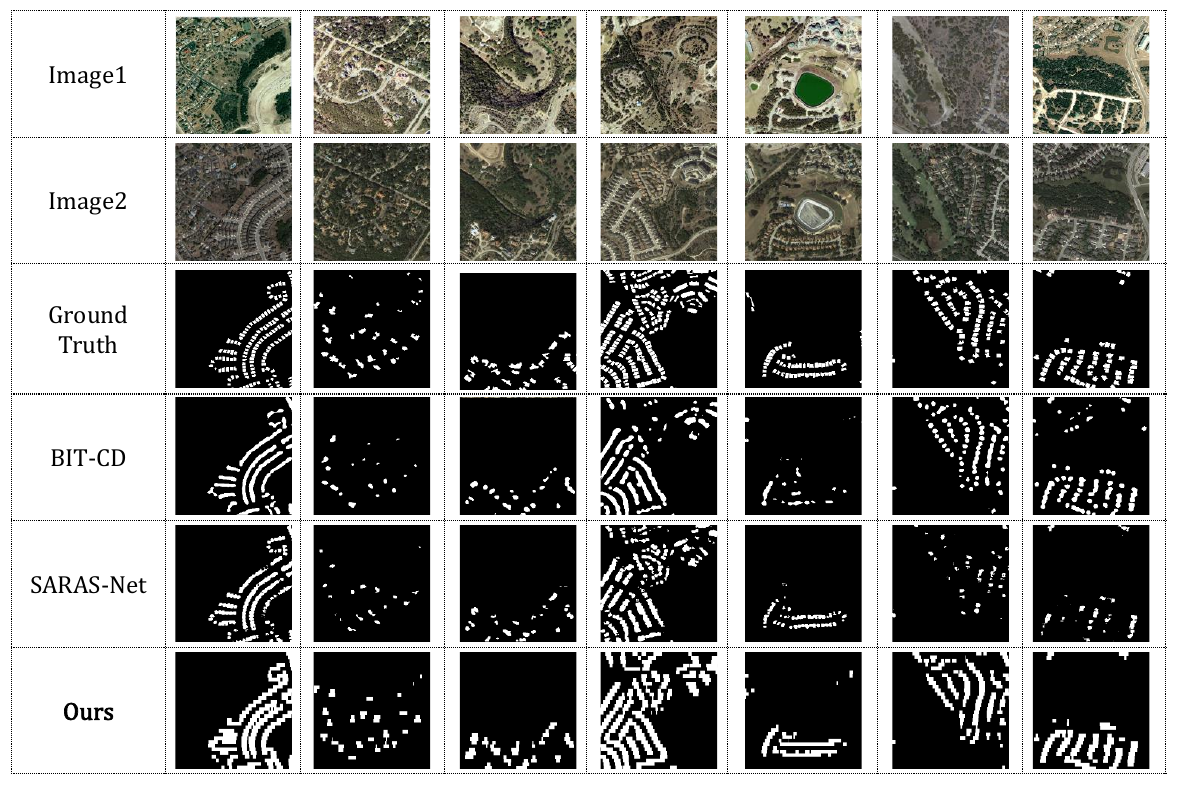}   
\caption{Qualitative results of different segmentation models on the LEVIR-CD dataset.}
\label{fig:LEVIR_result}
\end{figure}

Our proposed model also uses change detection methods. 
Since the main target of change detection is in the field of remote sensing, we also evaluate its performance on the publicly available remote sensing benchmark dataset.
We conduct comparative experiments using the LEVIR-CD dataset, a public remote sensing change detection dataset.
We train BIT-CD, SARAS-Net, and our TCD model on the LEVIR-CD train set from scratch for 200 epochs.
The test results are presented in the \cref{tab:seg_result_LEIVR} and \cref{fig:LEVIR_result}.
Experimental results show that BIT-CD and SARAS-Net produce fewer false positives, but they fail to detect many actual changes. 
In contrast, although this results in relatively more false positives, our model detected most of the changes and demonstrated the highest performance in F1 score, IoU, and OA. 
This is because our primary goal is to detect infrequently changed content within a large volume of text in the contract, leading us to prioritize improving recall, even at the expense of increased false positives. 

\begin{table}[t]
\centering
\resizebox{.50\columnwidth}{!}{
\begin{tabular}{l|c c c c c}
    \toprule
    Model                                           &  Pre.                     & Rec.                  & F1                    & IoU                   &  OA                   \\
    \midrule
    BIT-CD \cite{chen2021remote}                                          & 74.8                      & 71.9                  & 73.3                  & 57.9                  & 97.3                  \\
    SARAS-Net \cite{chen2023saras}                                       & \textbf{90.8}             & 62.9                  & 74.3                  & 59.1                  & \textbf{97.5}         \\
    \textbf{TCD(ours)}                              & 76.4                      & \textbf{77.6}         & \textbf{77.0}         & \textbf{62.6}         & 96.5                  \\
    \bottomrule
\end{tabular}
}
\caption{Quantitative results of the segmentation benchmark on the LEVIR-CD dataset.}
\label{tab:seg_result_LEIVR}
\end{table}

\subsection{Processing time for real contract documents}

We develop a DCD application based on our proposed TCD model to compare draft and scanned-signed versions of contract documents. 
To test the application, we create a test set consisting of real-world contracts.
Our dataset consists of 344 contracts, comprising 4044 pages of original documents and 4231 pages of scanned documents.
Of these, 72 contracts are bilingual, while the remaining are monolingual.
The contracts feature texts in multiple languages, including English, Vietnamese, Turkish, Hindi, Korean, Russian, Chinese, Latvian, Arabic, and Spanish.
Due to data confidentiality issues, we can only report numerical results without sharing the actual experimental samples.
We evaluate the performance of our TCD model on an NVIDIA RTX 2080 Ti, which utilize 2.7GB of memory and achieve an average processing time of 2.82 seconds per page.
This result includes the entire processing pipeline, from inputting the two documents to generating output after completing preprocessing, layout detection, unit detection, and the proposed TCD model.
We evaluate our approach using the entire dataset, and some of the evaluation results along with the average value of the whole data are presented in \cref{tab:speed_dcd}.

\begin{table}[t]
\centering
\resizebox{.75\columnwidth}{!}{
\begin{tabular}{l|l|c|c|c|c}
     \toprule
    No.    & language & page(org) & page(scan) & time(sec) & time/page(sec) \\
    \midrule
    1           & English,Spanish      & 73 & 73       & 303.76    & 2.08 \\
    2           & English,Hindi        & 8  & 8        & 47.81     & 2.99 \\
    3           & English,Turkish      & 4  & 4        & 19.26     & 2.41 \\
    4           & English,Latvian      & 10 & 15       & 60.03     & 2.40 \\
    5           & English,Russian      & 15 & 16       & 77.66     & 2.51 \\
    6           & Korean               & 5  & 5        & 16.54     & 1.65 \\
    7           & English              & 14 & 14       & 68.81     & 2.46 \\
    8           & Spanish              & 11 & 13       & 46.10     & 1.92 \\ 
    9           & Arabic               & 5  & 6        & 23.20     & 2.11 \\
    10          & Chinese              & 42 & 42       & 129.933   & 1.55 \\   
    \midrule
    Total(344)  & All                  & 4044 & 4231   & 25008.04 & \textbf{2.82}  \\
    \bottomrule
\end{tabular}
}
\caption{Quantitative results of document comparison time performance.}
\label{tab:speed_dcd}
\end{table}

\end{appendices}

\end{document}